\documentclass[11pt,a4paper]{article}
\usepackage[hyperref]{acl2021}
\usepackage{times}
\usepackage{latexsym}

\usepackage{microtype}
\usepackage{adjustbox}
\usepackage{booktabs}
\usepackage{multirow}
\usepackage{multicol}
\usepackage{breakurl}

\aclfinalcopy

\title{Re-Evaluating GermEval17 Using German Pre-Trained Language Models}

\author{Matthias~Aßenmacher$^1$\textsuperscript{$\spadesuit$} \\\And
  Alessandra Corvonato$^1$\textsuperscript{$\clubsuit$} \\ \\
  $^1$ Department of Statistics, Ludwig-Maximilians-Universität, Munich, Germany \\ \\
  \small\textsuperscript{$\spadesuit$}\texttt{\{matthias,chris\}@stat.uni-muenchen.de},\quad \textsuperscript{$\clubsuit$}\texttt{alessandracorvonato@yahoo.de} \\ \\ \textit{\footnotesize Accepted as a conference paper at the 6th Swiss Text AnalyticsConference (SwissText), Brugg, Switzerland (Online), June 14-16, 2021}  \\\And
  Christian Heumann$^1$\textsuperscript{$\spadesuit$}
}
\date{}

\begin{document}
\maketitle
\begin{abstract}
The lack of a commonly used benchmark data set (collection) such as (Super) GLUE \citep{wang2018glue, wang2019superglue} for the evaluation of non-English pre-trained language models is a severe shortcoming of current English-centric NLP-research. It concentrates a large part of the research on English, neglecting the uncertainty when transferring conclusions found for the English language to other languages. We evaluate the performance of German and multilingual BERT models currently available via the huggingface \texttt{transformers} library on four subtasks of Aspect-based Sentiment Analysis (ABSA) from the GermEval17 workshop. We compare them to pre-BERT architectures \citep{wojatzki2017germeval,schmitt-etal-2018-joint,attia-etal-2018-multilingual} as well as to an ELMo-based architecture \citep{biesialska2020sentiment} and a BERT-based approach \citep{guhr2020training}. The observed improvements are put in relation to those for a similar ABSA task \citep{semeval2014} and similar models (pre-BERT vs. BERT-based) for the English language and we check whether the reported improvements correspond to those we observe for German.
\end{abstract}

\section{Introduction}
\label{sec:intro}

(Aspect-based) Sentiment Analysis is often used to transform reviews into helpful information on how a product or service of a company is perceived among the customers. Until recently, Sentiment Analysis was mainly conducted using traditional machine learning and recurrent neural networks, like LSTMs \citep{hochreiter1997long} or GRUs \citep{cho2014learning}. Those models have been practically replaced by language models relying on (parts of) the Transformer architecture, a novel framework proposed by \citet{vaswani2017attention}. \citet{devlin2019bert} developed a Transformer-encoder-based language model called BERT (\textbf{B}idirectional \textbf{E}ncoder \textbf{R}epresentations from \textbf{T}ransfomers), achieving state-of-the-art (SOTA) performance on several benchmark tasks - mainly for the English language - and becoming a milestone in the field of NLP.

Up to now, only a few researchers have focused on sentiment related problems for German reviews, despite language-specific evaluation is a crucial driving force for a more universal model development and improvement. Unique characteristics of the different languages present different challenges to the models, which is why sole evaluation on English data is a severe shortcoming.

The first shared task on German ABSA, which provides a large annotated data set for training and evaluation, is the \textit{GermEval17 Shared Task} \citep{wojatzki2017germeval}. The participating teams back then analyzed the data using mostly standard machine learning techniques such as SVMs, CRFs, or LSTMs. In contrast to 2017, today, different pre-trained BERT models are available for a variety of different languages, including German. We re-analyzed the complete GermEval17 Task using seven pre-trained BERT models suitable for German provided by the huggingface \texttt{transformers} library \citep{Wolf2020HuggingFace}. We evaluate which one of the models is best suited for the different GermEval17 subtasks by comparing their performance values. Furthermore, we compare our findings on whether (and how much) BERT-based models are able to improve the pre-BERT SOTA in German ABSA with the SOTA developments for English ABSA by the example of SemEval-2014 \citep{semeval2014}.

We first give an overview on the GermEval17 tasks (cf. Sec. \ref{sec:germeval}) and on related work (cf. Sec. \ref{sec:related}). Second, we present the data and the models (cf. Sec. \ref{sec:models}), while Section \ref{sec:results} holds the results of our re-evaluation. Sections \ref{sec:disc} and \ref{sec:conc} conclude our work by stating our main findings and drawing parallels to the English language.

\section{The GermEval17 Task(s)}
\label{sec:germeval}

The GermEval17 Shared Task \citep{wojatzki2017germeval} is a task on analyzing aspect-based sentiments in customer reviews about "Deutsche Bahn" (DB) - the German public train company. The main data was crawled from various social media platforms such as Twitter, Facebook and Q\&A websites from May 2015 to June 2016. The documents were manually annotated, and split into a training (\textbf{train}), a development (\textbf{dev}) and a synchronic ($\textbf{test}_{syn}$) test set. A diachronic test set ($\textbf{test}_{dia}$) was collected the same way from November 2016 to January 2017 in order to test for temporal robustness. 
The task comprises four subtasks representing a complete classification pipeline. Subtask A is a binary Relevance Classification task which aims at identifying whether the feedback refers to DB. Subtask B aims at classifying the Document-level Polarity ("negative", "positive" and "neutral"). In Subtask C, the model has to identify all the aspect categories with associated sentiment polarities in a relevant document. This multi-label classification task was divided into Subtask C1 (\textit{Aspect-only}) and Subtask C2 (\textit{Aspect+Sentiment}). For this purpose, the organizers defined 20 different aspect categories, e.g. \texttt{Allgemein} (\textit{General}), \texttt{Sonstige Unregelmäßigkeiten} (\textit{Other irregularities}). Finally, Subtask D refers to the Opinion Target Extraction (OTE), i.e. a sequence labeling task extracting the linguistic phrase used to express an opinion. We differentiate between exact match (Subtask D1) and overlapping match, tolerating errors of $+/-$ one token (Subtask D2). 

\section{Related Work}
\label{sec:related}

Already before BERT, many researchers focused on (English) Sentiment Analysis \citep{behdenna2018document}. The most common architectures were traditional machine learning classifiers and recurrent neural networks (RNNs). SemEval14 \citep[Task 4;][]{semeval2014} was the first workshop to introduce Aspect-based Sentiment Analysis (ABSA) which was expanded within SemEval15 Task 12 \citep{semeval2015} and SemEval16 Task 5 \citep{semeval2016}. Here, restaurant and laptop reviews were examined on different granularities. The best model at SemEval16 was an SVM/CRF architecture using GloVe embeddings \citep{pennington2014glove}. 
However, many works recently focused on re-evaluating the SemEval Sentiment Analysis task using BERT-based language models \citep{hoang-etal-2019-aspect, xu2019bert, sun2019utilizing, li2019exploiting, karimi2020adversarial, tao2020toward}. 

In comparison, little research deals with German ABSA. For instance, \citet{barriere2020improving} trained a multilingual BERT model for German Document-level Sentiment Analysis on the SB-10k data set \citep{cieliebak2017}. Regarding the GermEval17 Subtask B, \citet{guhr2020training} considered both FastText \citep{bojanowski2017enriching} and BERT, achieving notable improvements. \citet{biesialska2020sentiment} made use of ensemble models: One is an ensemble of ELMo \citep{peters2018deep}, GloVe and a bi-attentive classification network \citep[BCN;][]{mccann2017learned}, achieving a score of 0.782, and the other one consists of ELMo and a Transformer-based Sentiment Analysis model (TSA), reaching a score of 0.789 for the synchronic test data set. Moreover, \citet{attia-etal-2018-multilingual} trained a convolutional neural network (CNN), achieving a score of 0.7545 on the synchronic test set. \citet{schmitt-etal-2018-joint} advanced the SOTA for Subtask C by employing biLSTMs and CNNs to carry out end-to-end Aspect-based Sentiment Analysis. The highest score was achieved using an end-to-end CNN architecture with FastText embeddings, scoring 0.523 and 0.557 on the synchronic and diachronic test data set for Subtask C1, respectively, and 0.423 and 0.465 for Subtask C2. 

\section{Materials and Methods}
\label{sec:models}

\paragraph{Data}

The GermEval17 data is freely available in \texttt{.xml}- and \texttt{.tsv}-format\footnote{The data sets (in both formats) can be obtained from \href{http://ltdata1.informatik.uni-hamburg.de/germeval2017/}{http://ltdata1.informatik.uni-hamburg.de/germeval2017/}.}. 
Each data split (train, validation, test) in \texttt{.tsv}-format contains the following variables: 
 \vspace{-.15cm}
\begin{itemize}
    \item document id (URL) \vspace{-.3cm}
    \item document text \vspace{-.3cm}
    \item relevance label (\texttt{true}, \texttt{false})
    \item document-level sentiment label\\{\small(\texttt{negative},  \texttt{neutral}, \texttt{positive})} \vspace{-.3cm}
    \item aspects with respective polarities\\{\small(e.g. \texttt{Ticketkauf\#Haupt:negative})} \vspace{-.3cm}
\end{itemize}

For documents which are annotated as irrelevant, the sentiment label is set to \texttt{neutral} and no aspects are available. Visibly, the \texttt{.tsv}-formatted data does not contain the target expressions or their associated sequence positions. Consequently, Subtask D can only be conducted using the data in \texttt{.xml}-format, which additionally holds the information on the starting and ending sequence positions of the target phrases.

The data set comprises $\sim 26$k documents in total, including the diachronic test set with around $1.8$k examples. Further, the main data was randomly split by the organizers into a train data set for training, a development data set for validation and a synchronic test data set.
Table \ref{tab:docs_per_df} displays the number of documents for each split.

\begin{table}[ht]
    \centering
    \begin{adjustbox}{width=.35\textwidth}
    \begin{tabular}{|rrrr|}
    \toprule
        \textbf{train} & \textbf{dev} & $\textbf{test}_{syn}$ & $\textbf{test}_{dia}$ \\
    \midrule
        19,432 & 2,369 & 2,566 & 1,842 \\
    \bottomrule
    \end{tabular}
    \end{adjustbox}
    \caption{Number of documents per split of the data set.}
    \label{tab:docs_per_df}
\end{table}

\noindent While roughly 74\% of the documents form the train set, the development split and the synchronic test split contain around 9\% and around 10\%, respectively. The remaining 7\% of the data belong to the diachronic set (cf. Tab. \ref{tab:docs_per_df}). Table \ref{tab:relevance_distribution} shows the relevance distribution per data split. This unveils a pretty skewed distribution of the labels since the relevant documents represent the clear majority with over 80\% in each split.

\begin{table}[ht]
    \centering
    \begin{adjustbox}{width=.45\textwidth}
    \begin{tabular}{|l|rrrr|}
    \toprule
        \textbf{Relevance} & \textbf{train} & \textbf{dev} & $\textbf{test}_{syn}$ & $\textbf{test}_{dia}$ \\
    \midrule
        true & 16,201 & 1,931 & 2,095 & 1,547 \\
        false & 3,231 & 438 & 471 & 295 \\
    \bottomrule
    \end{tabular}
    \end{adjustbox}
    \caption{Relevance distribution for Subtask A.}
    \label{tab:relevance_distribution}
\end{table}

\noindent The distribution of the sentiments is depicted in Table \ref{tab:sentiment_distribution}, which shows that between 65\% and 69\% (per split) belong to the neutral class, 25--31\% to the negative and only 4--6\% to the positive class.

Table \ref{tab:distribution_aspects} holds the distribution of the 20 different aspect categories assigned to the documents\footnote{Multiple annotations per document are possible; for a detailed category description see  \href{https://sites.google.com/view/germeval2017-absa/data}{https://sites.google.com/view/germeval2017-absa/data}.}. It shows the number of documents containing certain categories without differentiating between how often a category appears within a given document.

\begin{table}[ht]
    \centering
    \begin{adjustbox}{width=.45\textwidth}
    \begin{tabular}{|l|rrrr|}
    \toprule
        \textbf{Sentiment} & \textbf{train} & \textbf{dev} & $\textbf{test}_{syn}$ & $\textbf{test}_{dia}$ \\
    \midrule
        negative & 5,045 & 589 & 780 & 497 \\
        neutral & 13,208 & 1,632 & 1,681 & 1,237 \\
        positive & 1,179 & 148 & 105 & 108 \\
    \bottomrule
    \end{tabular}
    \end{adjustbox}
    \caption{Sentiment distribution for Subtask B.}
    \label{tab:sentiment_distribution}
\end{table}

\noindent The relative distribution of the aspect categories is similar between the splits. On average, there are $\sim 1.12$ different aspects per document. Again, the label distribution is heavily skewed, with \texttt{Allgemein} (\textit{General}) clearly representing the majority class, as it is present in 75.8\% of the documents with aspects.
The second most frequent category is \texttt{Zugfahrt} (\textit{Train ride}) appearing in around 13.8\% of the documents. This strong imbalance in the aspect categories leads to an almost Zipfian distribution \citep{wojatzki2017germeval}. \\

\begin{table}[ht]
    \centering
    \begin{adjustbox}{width=.49\textwidth}
    \begin{tabular}{|l|rrrr|}
    \toprule
    \textbf{Category} & \textbf{train} & \textbf{dev} & $\textbf{test}_{syn}$ & $\textbf{test}_{dia}$ \\
    \midrule
Allgemein & 11,454 & 1,391 & 1,398 & 1,024 \\
Zugfahrt & 1,687 & 177 & 241 & 184 \\
Sonstige Unregelmäßigkeiten & 1,277 & 139 & 224 & 164 \\
Atmosphäre & 990 & 128 & 148 & 53 \\
Ticketkauf & 540 & 64 & 95 & 48 \\
Service und Kundenbetreuung & 447 & 42 & 63 & 27 \\
Sicherheit & 405 & 59 & 84 & 42 \\
Informationen & 306 & 28 & 58 & 35 \\
Connectivity & 250 & 22 & 36 & 73 \\
Auslastung und Platzangebot & 231 & 25 & 35 & 20 \\
DB App und Website & 175 & 20 & 28 & 18 \\
Komfort und Ausstattung & 125 & 18 & 24 & 11 \\
Barrierefreiheit & 53 & 14 & 9 & 2 \\
Image & 42 & 6 & 0 & 3 \\
Toiletten & 41 & 5 & 7 & 4 \\
Gastronomisches Angebot & 38 & 2 & 3 & 3 \\
Reisen mit Kindern & 35 & 3 & 7 & 2 \\
Design & 29 & 3 & 4 & 2 \\
Gepäck & 12 & 2 & 2 & 6 \\
QR-Code & 0 & 1 & 1 & 0 \\
\midrule
total & 18,137 & 2,149 & 2,467 & 1,721 \\
\# documents with aspects & 16,200 & 1,930 & 2,095 & 1,547 \\
$\emptyset$ different aspects/document & 1.12 & 1.11 & 1.18 & 1.11 \\
    \bottomrule
    \end{tabular}
    \end{adjustbox}
    \caption{Aspect category distribution for Subtask C. Multiple mentions of the same aspect category in a document are only considered once.}
    \label{tab:distribution_aspects}
\end{table}

\begin{table*}[ht]
    \centering
    \begin{adjustbox}{width=\textwidth}
    \begin{tabular}{|l|l|l|}
    \toprule
        Model variant & Pre-training corpus & Properties \\
    \midrule
        \texttt{bert-base-german-cased} & 12GB of German text (deepset.ai) & L=12, H=768, A=12, 110M parameters \\
        \texttt{bert-base-german-dbmdz-cased} & 16GB of German text (dbmdz) & L=12, H=768, A=12, 110M parameters \\
        \texttt{bert-base-german-dbmdz-uncased} & 16GB of German text (dbmdz) & L=12, H=768, A=12, 110M parameters \\
        \texttt{bert-base-multilingual-cased} & Largest Wikipedias (top 104 languages) & L=12, H=768, A=12, 179M parameters \\
        \texttt{bert-base-multilingual-uncased} & Largest Wikipedias  (top 102 languages) & L=12, H=768, A=12, 168M parameters \\
        \texttt{distilbert-base-german-cased} & 16GB of German text (dbmdz) & L=6, H=768, A=12, 66M parameters \\
       \texttt{distilbert-base-multilingual-cased} & Largest Wikipedias (top 104 languages) & L=6, H=768, A=12, 134M parameters \\
        \bottomrule
    \end{tabular}
    \end{adjustbox}
    \caption{Pre-trained models provided by huggingface \texttt{transformers} (version 4.0.1) suitable for German. For \textbf{all} available models, see: \url{https://huggingface.co/transformers/pretrained_models.html}.}
    \label{tab:LMs}
\end{table*}

\paragraph{Pre-trained architectures} BERT was initially introduced in a \texttt{base} (110M parameters) and a \texttt{large} (340M) variant, \citet{sanh2019distilbert} proposed an even smaller BERT model (DistilBERT, 60M parameters) trained via knowledge distillation \citep{hinton2015distilling}. The exact model specifications regarding number of layers ($L$), number of attention heads ($A$) and embedding size ($H$) for available German BERT models are depicted in the last column of Table \ref{tab:LMs}. Both architectures were pre-trained on the Masked Language Modeling task as well as on the auxiliary Next Sentence Prediction task (only BERT) and can subsequently be fine-tuned on a task at hand.

We include three German (Distil)BERT models pre-trained by DBMDZ\footnote{MDZ Digital Library team at the Bavarian State Library. Visit \href{https://www.digitale-sammlungen.de}{https://www.digitale-sammlungen.de} for details and \href{https://github.com/dbmdz/berts}{https://github.com/dbmdz/berts} for their repository on pre-trained BERT models.} and one by Deepset.ai\footnote{Visit \href{https://deepset.ai/german-bert}{https://deepset.ai/german-bert} for details.}. 
The latter one is pre-trained using German \texttt{Wikipedia} (6GB raw text files), the \texttt{Open Legal Data} dump \citep[2.4GB;][]{openlegaldata} and news articles (3.6GB). 
DBMDZ combined \texttt{Wikipedia}, \texttt{EU Bookshop} \citep{eubookshop}, \texttt{Open Subtitles} \citep{opensubtitles}, \texttt{CommonCrawl} \citep{commoncrawl}, \texttt{ParaCrawl} \citep{paracrawl} and \texttt{News Crawl} \citep{newscrawl} to a corpus with a total size of 16GB with $\sim 2,350$M tokens. Besides this, we use the three multilingual (Distil)BERT models included in the \texttt{transformers} module. This amounts to five BERT and two DistilBERT models, two of which are "uncased" (i.e. every character is lower-cased) while the other five models are "cased" ones.

\section{Results}
\label{sec:results}

For the re-evaluation, we used the latest data provided in \texttt{.xml}-format. Duplicates were not removed, in order to make our results as comparable as possible. We tokenized the documents and fixed single spelling mistakes in the labels\footnote{"positve" in \textbf{train} set was replaced with "positive",\\ " ~negative" in $\textbf{test}_{dia}$ set was replaced with "negative".}. For Subtask D, the BIO-tags were added based on the provided sequence positions, i.e. one entity corresponds to at least one token tag starting with \texttt{B-} for "Beginning" and continuing with \texttt{I-} for "Inner". If a token does not belong to any entity, the tag \texttt{O} for "Outer" is assigned. For instance, the sequence "fährt nicht" (engl. "does not run") consists of two tokens and would receive the entity \texttt{Zugfahrt:negative} and the token tags [\texttt{B-Zugfahrt:negative},  \texttt{I-Zugfahrt:negative}] if it refers to a DB train which is not running.

The models were fine-tuned on one Tesla V100 PCIe 16GB GPU using Python 3.8.7. Moreover, the \texttt{transformers} module (version 4.0.1) and \texttt{torch} (version 1.7.1) were used\footnote{Source code is available on GitHub: \href{https://github.com/ac74/reevaluating_germeval2017}{https://github.com/ac74/reevaluating\_germeval2017}. 
The results are fully reproducible for Subtasks A, B and C. For Subtask D, reproducibility could not be ensured. The micro F1 scores fluctuate across different runs between +/-0.01 around the reported values.}.
The considered values for the hyperparameters for fine-tuning follow the recommendations of \citet{devlin2019bert}:

\begin{itemize}
\vspace{-.3cm}    
    \item Batch size $\in \{16, 32\}$, \vspace{-.3cm}
    \item Adam learning rate $\in \{\mbox{5e,3e,2e}\}-5$, \vspace{-.3cm}
    \item \# epochs $\in \{2, 3, 4\}$. \vspace{-.3cm}
\end{itemize}

After evaluating the model performance for combinations\footnote{Due to memory limitations, not every hyperparameter combination was applicable.} of the different hyperparameters, all pre-trained architectures were fine-tuned with a learning rate of $\mbox{5e-5}$ for four epochs, which turned out to be the most promising combination across the different models. The maximum sequence length was set to 256, which is sufficient since the evaluated data set consists of rather short texts from social media, and a batch size of 32 was chosen.

\paragraph{Other models}

Eight teams officially participated in the GermEval17 shared task, five of which analyzed Subtask A, all of them Subtask B and two repectively Subtask C and D. We furthermore consider the system by \citet{LT-ABSA} additionally to the participants' models from 2017, even though they were the organizers and did not "officially" participate. They also tackled all four subtasks. Since 2017 several other authors analyzed (parts of) the GermEval17 subtasks using more advanced models, which we also consider for comparison here. Table \ref{tab:overview} shows which authors employed which kinds of models to solve which task. 

\begin{table}[ht]
    \centering
    \begin{adjustbox}{width=.49\textwidth}
    \begin{tabular}{|l||c|c|cc|cc|}
    \toprule
        \textbf{Subtask} & A & B & C1 & C2 & D1 & D2  \\
        \midrule
        Models from 2017 & \multirow{2}{*}{X} & \multirow{2}{*}{X} & \multirow{2}{*}{X} & \multirow{2}{*}{X} & \multirow{2}{*}{X} & \multirow{2}{*}{X}  \\
        \citep{wojatzki2017germeval, LT-ABSA} &  &  &  &  &  &  \\
        \midrule
	Our BERT models	& X & X & X & X & X & X 		\\
        \midrule
    CNN \citep{attia-etal-2018-multilingual} & -- & X & -- & -- & -- & -- \\
    CNN+FastText \citep{schmitt-etal-2018-joint} & -- & -- & X & X & -- & -- \\
    ELMo+GloVe+BCN \citep{biesialska2020sentiment} & -- & X & -- & -- & -- & -- \\
    ELMo+TSA \citep{biesialska2020sentiment} & -- & X & -- & -- & -- & -- \\
    FastText \citep{guhr2020training} & -- & X & -- & -- & -- & -- \\
    \texttt{bert-base-german-cased} & \multirow{2}{*}{--} & \multirow{2}{*}{X} & \multirow{2}{*}{--} & \multirow{2}{*}{--} & \multirow{2}{*}{--} & \multirow{2}{*}{--} \\
    \citep{guhr2020training} &  &  &  &  &  &  \\
        \bottomrule
    \end{tabular}
    \end{adjustbox}
    \caption{An overview on all the models discussed in this article, an "X" in a column indicates that the architecture was evaluated on the respective subtask.}
    \label{tab:overview}
\end{table}

\paragraph{Subtask A}

The Relevance Classification is a binary document classification task with classes \texttt{true} and \texttt{false}. Table \ref{tab:myresultsA} displays the micro F1 score obtained by each language model on each test set (best result per data set in bold).

\begin{table}[ht]
    \centering
    \begin{adjustbox}{width=.49\textwidth}
    \begin{tabular}{|l||c|c|}
    \toprule
        \textbf{Language model} &$\textbf{test}_{syn}$ & $\textbf{test}_{dia}$  \\
        \midrule
        Best model 2017 \citep{IDS_IUCL} & 0.903 & 0.906 \\
        \midrule
	\texttt{bert-base-german-cased}	&	0.950	&	0.939		\\
    \texttt{bert-base-german-dbmdz-cased}	&	0.951	&	0.946\\
    \texttt{bert-base-german-dbmdz-uncased}	&	\textbf{0.957}	&	\textbf{0.948}\\
	\texttt{bert-base-multilingual-cased}	&	0.942	&	0.933		\\
	\texttt{bert-base-multilingual-uncased}	&	0.944	&	0.939		\\
	\texttt{distilbert-base-german-cased}	&	0.944	&	0.939	\\
	\texttt{distilbert-base-multilingual-cased}	&	0.941	&	0.932\\
        \bottomrule
    \end{tabular}
    \end{adjustbox}
    \caption{F1 scores for Subtask A on synchronic and diachronic test sets.}
    \label{tab:myresultsA}
\end{table}

\noindent All the models outperform the best result achieved in 2017 for both test data sets. For the synchronic test set, the previous best result is surpassed by 3.8--5.4 percentage points. For the diachronic test set, the absolute difference to the best contender of 2017 varies between 2.6 and 4.2 percentage points. With a micro F1 score of 0.957 and 0.948, respectively, the best scoring pre-trained language model is the uncased German BERT-BASE variant by \texttt{dbmdz}, followed by its cased version. All the pre-trained models perform slightly better on the synchronic test data than on the diachronic data. \citet{attia-etal-2018-multilingual}, \citet{schmitt-etal-2018-joint}, \citet{biesialska2020sentiment} and \citet{guhr2020training} did not evaluate their models on this task.

\paragraph{Subtask B}

Subtask B refers to the Document-level Polarity, which is a multi-class classification task with three classes. Table \ref{tab:myresultsB} demonstrates the performances on the two test sets:

\begin{table}[ht]
    \centering
    \begin{adjustbox}{width=.49\textwidth}
    \begin{tabular}{|l||c|c|}
    \toprule
        \textbf{Language model} &$\textbf{test}_{syn}$ & $\textbf{test}_{dia}$  \\
        \midrule
        Best models 2017 \citep[\textbf{test}$_{syn}$:][]{LT-ABSA} & \multirow{2}{*}{0.767} & \multirow{2}{*}{0.750} \\
        \citep[\textbf{test}$_{dia}$:][]{IDS_IUCL} &  &  \\
        \midrule
	\texttt{bert-base-german-cased}	&	0.798		&	0.793	\\
    \texttt{bert-base-german-dbmdz-cased}	&	0.799	&	0.785	\\
    \texttt{bert-base-german-dbmdz-uncased}	&	\textbf{0.807}	&	\textbf{0.800}\\
	\texttt{bert-base-multilingual-cased}	&	0.790		&	0.780	\\
	\texttt{bert-base-multilingual-uncased} &	0.784		&	0.766		\\
	\texttt{distilbert-base-german-cased}	&	0.798		&	0.776		\\
	\texttt{distilbert-base-multilingual-cased} 	&	0.777	&	0.770	\\
        \midrule
    CNN \citep{attia-etal-2018-multilingual} & 0.755 & -- \\
    ELMo+GloVe+BCN \citep{biesialska2020sentiment} & 0.782 & -- \\
    ELMo+TSA \citep{biesialska2020sentiment} & 0.789 & -- \\
    FastText \citep{guhr2020training} & 0.698\textsuperscript{\textdagger} & -- \\
    \texttt{bert-base-german-cased} \citep{guhr2020training} & 0.789\textsuperscript{\textdagger} & -- \\
        \bottomrule
    \end{tabular}
    \end{adjustbox}
    \caption{Micro-averaged F1 scores for Subtask B on synchronic and diachronic test sets.
    \newline \textsuperscript{\textdagger}\textit{\citet{guhr2020training} created their own (balanced \& unbalanced) data splits, which limits comparability. We compare to the performance on the unbalanced data since it more likely resembles the original data splits.}}
    \label{tab:myresultsB}
\end{table}

\noindent All models outperform the best model from 2017 by 1.0--4.0 percentage points for the synchronic, and by 1.6--5.0 percentage points for the diachronic test set. On the synchronic test set, the uncased German BERT-BASE model by \texttt{dbmdz} performs best with a score of 0.807, followed by its cased variant with 0.799.
For the diachronic test set, the uncased German BERT-BASE model exceeds the other models with a score of 0.800, followed by the cased German BERT-BASE model reaching a score of 0.793. The three multilingual models perform generally worse than the German models on this task. Besides this, all the models perform slightly better on the synchronic data set than on the diachronic one. The FastText-based model \citep{guhr2020training} comes not even close to the baseline from 2017, while the ELMo-based models \citep{biesialska2020sentiment} are pretty competitive. Interestingly, two of the multilingual models are even outperformed by these ELMo-based models.

\paragraph{Subtask C}

Subtask C is split into \textit{Aspect-only} (Subtask C1) and \textit{Aspect+Sentiment} Classification (Subtask C2), each being a multi-label classification task\footnote{This leads to a change of activation functions in the final layer from softmax to sigmoid + binary cross entropy loss.}. As the organizers provide 20 aspect categories, Subtask C1 includes 20 labels, whereas Subtask C2 has 60 labels since each aspect category can be combined with each of the three sentiments. Consistent with \citet{UKP_Lab_TUDA} and \citet{im_sing}, we do not account for multiple mentions of the same label in one document.
The results for Subtask C1 are shown in Table \ref{tab:myresultsC1}:

\begin{table}[ht]
    \centering
    \begin{adjustbox}{width=.49\textwidth}
    \begin{tabular}{|l||c|c|}
    \toprule
        \textbf{Language model} &$\textbf{test}_{syn}$ & $\textbf{test}_{dia}$  \\
        \midrule
        Best model 2017 \citep{LT-ABSA} & 0.537  & 0.556 \\
        \midrule
	\texttt{bert-base-german-cased}	&	0.756	&	0.762		\\
    \texttt{bert-base-german-dbmdz-cased}	&	0.756	&	0.781	\\
    \texttt{bert-base-german-dbmdz-uncased}	&	\textbf{0.761}	&	\textbf{0.791}		\\
	\texttt{bert-base-multilingual-cased}	&	0.706	&	0.734	\\
	\texttt{bert-base-multilingual-uncased}	&	0.723	&	0.752	\\
	\texttt{distilbert-base-german-cased}	&	0.738	&	0.768	\\
    \texttt{distilbert-base-multilingual-cased}	&	0.716	&	0.744	\\
        \midrule
    CNN+FastText \citep{schmitt-etal-2018-joint} & 0.523 & 0.557 \\
        \bottomrule
    \end{tabular}
    \end{adjustbox}
    \caption{Micro-averaged F1 scores for Subtask C1 (\textit{Aspect-only}) on synchronic and diachronic test sets. A detailed overview of \textit{per-class} performances for error analysis can be found in Table \ref{tab:detail_C1} in Appendix \ref{a:details-c}.}
    \label{tab:myresultsC1}
\end{table}

\noindent All pre-trained German BERTs clearly surpass the best performance from 2017 as well as the results reported by \citet{schmitt-etal-2018-joint}, who are the only ones of the other authors to evaluate their models on this tasks. Regarding the synchronic test set, the absolute improvement ranges between 16.9 and 22.4 percentage points, while for the diachronic test data, the models outperform the previous results by 17.8--23.5 percentage points. The best model is again the uncased German BERT-BASE model by \texttt{dbmdz}, reaching scores of 0.761 and 0.791, respectively, followed by the two cased German BERT-BASE models. One more time, the multilingual models exhibit the poorest performances amongst the evaluated models. Next, Table \ref{tab:myresultsC2} shows the results for Subtask C2:

\begin{table}[ht]
    \centering
    \begin{adjustbox}{width=.49\textwidth}
    \begin{tabular}{|l||c|c|}
    \toprule
        \textbf{Language model} &$\textbf{test}_{syn}$ & $\textbf{test}_{dia}$  \\
        \midrule
        Best model 2017 \citep{LT-ABSA} & 0.396 & 0.424\\
        \midrule
	\texttt{bert-base-german-cased}	&	0.634	&	0.663	\\
    \texttt{bert-base-german-dbmdz-cased}	&	0.628	&	0.663	\\
    \texttt{bert-base-german-dbmdz-uncased}	&	\textbf{0.655}	&	\textbf{0.689}		\\
	\texttt{bert-base-multilingual-cased}	&	0.571	&	0.634	\\
	\texttt{bert-base-multilingual-uncased}	&	0.553	&	0.631	\\
	\texttt{distilbert-base-german-cased}	&	0.629	&	0.663	\\
	\texttt{distilbert-base-multilingual-cased}	&	0.589	&	0.642	\\
        \midrule
    CNN+FastText \citep{schmitt-etal-2018-joint} & 0.423 & 0.465 \\
        \bottomrule
    \end{tabular}
    \end{adjustbox}
    \caption{Micro-averaged F1 scores for Subtask C2 (\textit{Aspect+Sentiment}) on synchronic and diachronic test sets. A detailed overview of \textit{per-class} performances for error analysis can be found in Table \ref{tab:detail_C2} in Appendix \ref{a:details-c}.}
    \label{tab:myresultsC2}
\end{table}

\noindent Here, the pre-trained models surpass the best model from 2017 by 15.7--25.9 percentage points and 20.7--26.5 percentage points, respectively, for the synchronic and diachronic test sets. Again, the best model is the uncased German BERT-BASE \texttt{dbmdz} model reaching scores of 0.655 and 0.689, respectively. The CNN models \citep{schmitt-etal-2018-joint} are also outperformed. For both, Subtask C1 and C2, all the displayed models perform better on the diachronic than on the synchronic test data. 

\paragraph{Subtask D}

Subtask D refers to the Opinion Target Extraction (OTE) and is thus a token-level classification task. As this is a rather difficult task, \citet{wojatzki2017germeval} distinguish between exact (Subtask D1) and overlapping match (Subtask D2), tolerating a deviation of $+/-$ one token. Here, "entities" are identified by their BIO-tags.
It is noteworthy that there are less entities here than for Subtask C since document-level aspects or sentiments could not always be assigned to a certain sequence in the document. As a result, there are less documents at disposal for this task, namely 9,193. The remaining data has 1.86 opinions per document on average. The majority class is now \texttt{Sonstige Unregelmäßigkeiten:negative} with around 15.4\% of the true entities (16,650 in total), leading to more balanced data than in Subtask C. 

\begin{table}[ht]
    \centering
    \begin{adjustbox}{width=.49\textwidth}
    \begin{tabular}{|c|l||c|c|}
    \toprule
        & \textbf{Language model} &$\textbf{test}_{syn}$ & $\textbf{test}_{dia}$  \\
        \midrule
        & Best model 2017 \citep{LT-ABSA} & 0.229 & 0.301 \\
        \midrule
	&	\texttt{bert-base-german-cased}	&	0.460	&	0.455	\\
\multirow{3}{*}{\rotatebox[origin=c]{90}{without CRF}} 
	&	\texttt{bert-base-german-dbmdz-cased}	&	0.480	&	0.466	\\
	&	\texttt{bert-base-german-dbmdz-uncased}	&	0.492	&	0.501		\\
	&	\texttt{bert-base-multilingual-cased}	&	0.447	&	0.457	\\
	&	\texttt{bert-base-multilingual-uncased}	&	0.429	&	0.404	\\
	&	\texttt{distilbert-base-german-cased}	&	0.347	&	0.357	\\
	&	\texttt{distilbert-base-multilingual-cased}	&	0.430	&	0.419	\\
	\midrule
	&	\texttt{bert-base-german-cased}	&	0.446	&	0.443	\\
	&	\texttt{bert-base-german-dbmdz-cased}	&	0.466	&	0.444	\\
\multirow{3}{*}{\rotatebox[origin=c]{90}{with CRF}} 
	&	\texttt{bert-base-german-dbmdz-uncased}	&	\textbf{0.515}	&	\textbf{0.518}	\\
	&	\texttt{bert-base-multilingual-cased}	&	0.472		&	0.466	\\
	&	\texttt{bert-base-multilingual-uncased}	&	0.477	&	0.452	\\
 	&	\texttt{distilbert-base-german-cased}	&	0.424		&	0.403	\\
	&	\texttt{distilbert-base-multilingual-cased}	&	0.436		&	0.418	\\
        \bottomrule
    \end{tabular}
    \end{adjustbox}
    \caption{Entity-level micro-averaged F1 scores for Subtask D1 (\textit{exact match}) on synchronic and diachronic test sets. A detailed overview of \textit{per-class} performances for error analysis can be found in Table \ref{tab:detail_D1} in Appendix \ref{a:details-d}.}
    \label{tab:myresultsD1}
\end{table}
\begin{table}[ht]
    \centering
    \begin{adjustbox}{width=.49\textwidth}
    \begin{tabular}{|c|l||c|c|}
    \toprule
        & \textbf{Language model} &\textbf{test}$_{syn}$ & \textbf{test}$_{dia}$  \\
        \midrule
        & Best models 2017 \citep[\textbf{test}$_{syn}$:][]{UKP_Lab_TUDA} & \multirow{2}{*}{0.348} & \multirow{2}{*}{0.365} \\
        & \citep[\textbf{test}$_{dia}$:][]{LT-ABSA} &&\\
        \midrule
	&	\texttt{bert-base-german-cased}	&	0.471	&	0.474	\\
\multirow{3}{*}{\rotatebox[origin=c]{90}{without CRF}} 
	&	\texttt{bert-base-german-dbmdz-cased}	&	0.491	&	0.488	\\
	&	\texttt{bert-base-german-dbmdz-uncased}	&	0.501	&	0.518	\\
	&	\texttt{bert-base-multilingual-cased}	&	0.457	&	0.473	\\
	&	\texttt{bert-base-multilingual-uncased}	&	0.435	&	0.417	\\
	&	\texttt{distilbert-base-german-cased}	&	0.397	&	0.407	\\
	&	\texttt{distilbert-base-multilingual-cased}	&	0.433	&	0.429	\\
	\midrule
	&	\texttt{bert-base-german-cased}	&	0.455	&	0.457		\\
	&	\texttt{bert-base-german-dbmdz-cased}	&	0.476		&	0.469	\\
\multirow{3}{*}{\rotatebox[origin=c]{90}{with CRF}} 
	&	\texttt{bert-base-german-dbmdz-uncased}	&	\textbf{0.523}		&	\textbf{0.533}	\\
	&	\texttt{bert-base-multilingual-cased}	&	0.476		&	0.474	\\
	&	\texttt{bert-base-multilingual-uncased}	&	0.484	&	0.464	\\
	&	\texttt{distilbert-base-german-cased}	&	0.433	&	0.423	\\
	&	\texttt{distilbert-base-multilingual-cased}	&	0.442	&	0.427	\\
        \bottomrule
    \end{tabular}
    \end{adjustbox}
    \caption{Entity-level micro-averaged F1 scores for Subtask D2 (\textit{overlapping match}) on synchronic and diachronic test sets. A detailed overview of \textit{per-class} performances for error analysis can be found in Table \ref{tab:detail_D2} in Appendix \ref{a:details-d}.}
    \label{tab:myresultsD2}
\end{table}

\noindent In Table \ref{tab:myresultsD1}, we compare the pre-trained models using an "ordinary" softmax layer to when using a CRF layer for Subtask D1.

The best performing model is the uncased German BERT-BASE model by \texttt{dbmdz} with CRF layer on both test sets, with a score of 0.515 and 0.518, respectively. Overall, the results from 2017 are outperformed by 11.8--28.6 percentage points on the synchronic test set and 5.6--21.7 percentage points on the diachronic test set. 

For the overlapping match (cf. Tab. \ref{tab:myresultsD2}), the best system from 2017 are outperformed by 4.9--17.5 percentage points on the synchronic and by 4.2--16.8 percentage points on the diachronic test set. Again, the uncased German BERT-BASE model by \texttt{dbmdz} with CRF layer performs best with an micro F1 score of 0.523 on the synchronic and 0.533 on the diachronic set. To our knowledge, there were no other models to compare our performance values with, besides the results from 2017.

\paragraph{Main Takeaways} 

For the first two subtasks, which are rather simple binary and multi-class classification tasks, the pre-trained models are able to improve a little upon the already pretty decent performance values from 2017. Further, we do not see large differences between the different pre-trained models. Nevertheless, the small differences we can observe, already point in the same direction as what can be observed for the primary ABSA tasks of interest, C1 and C2:

\begin{itemize}
\vspace{-.3cm}
    \item Uncased models have a tendency of outperforming their cased counterparts for the monolingual models, for multilingual models this cannot be clearly confirmed. \vspace{-.3cm}
    \item Monolingual models outperform the multilingual ones. \vspace{-.3cm}
    \item There are no large performance differences between the two cased BERT models by DBMDZ and Deepset.ai, which suggests only a minor influence of the different corpora, which the models were pre-trained on. \vspace{-.3cm}
    \item The monolingual DistilBERT model is pretty competitive, it consistently outperforms its multilingual counterpart as well as the multilingual BERT models on the subtasks A -- C and is at least competitive to the monolingual BERT models. \vspace{-.3cm}
\end{itemize}

For D1 and D2 we observe a rather clear dominance of the uncased monolingual model which is not observable to this extent for the other tasks.

\section{Discussion}
\label{sec:disc}

After having observed a notable performance increase for German ABSA when employing pre-trained models, the next step is to compare these observations to what was reported for the English language. 
Therefore, we examine the temporal development of the SOTA performance on the most widely adopted data sets for English ABSA, originating from the SemEval Shared Tasks \citep{semeval2014,semeval2015,semeval2016}. When looking at public leaderboards, e.g. \href{https://paperswithcode.com/}{https://paperswithcode.com/}, Subtask SB2 (\textit{aspect term polarity}) from SemEval-2014 is the task which attracts most of the researchers. This task is related, but not perfectly similar, to Subtask C2, since in this case, the \textit{aspect term} is always a word which has to present in the given review. For this task, a comparison of pre-BERT and BERT-based methods reveals no big "jump" in the performance values, but rather a steady increase over time (cf. Tab. \ref{tab:semeval14-sb2}).

\begin{table}[ht]
    \centering
    \begin{adjustbox}{width=.49\textwidth}
    \begin{tabular}{|c|l||c|c|}
    \toprule
        & \textbf{Language model} & Laptops & Restaurants  \\
        \midrule
        & Best model SemEval-2014 & \multirow{2}{*}{0.7048} & \multirow{2}{*}{0.8095} \\
        \multirow{3}{*}{\rotatebox[origin=c]{90}{pre-BERT}} & \citep{semeval2014} &&\\
    	&	\multirow{2}{*}{MemNet \citep{tang2016aspect}}	&	\multirow{2}{*}{0.7221}	&	\multirow{2}{*}{0.8095}	\\ &&&\\
    	&	\multirow{2}{*}{HAPN \citep{li-etal-2018-hierarchical}}	&	\multirow{2}{*}{0.7727}	&	\multirow{2}{*}{0.8223}	\\ &&&\\
	\midrule
	    &	\multirow{2}{*}{BERT-SPC \citep{song2019attentional}}	&	\multirow{2}{*}{0.7899}	&	\multirow{2}{*}{0.8446}	\\ \multirow{3}{*}{\rotatebox[origin=c]{90}{BERT-based}} &&&\\
    	&	\multirow{2}{*}{BERT-ADA \citep{rietzler-etal-2020-adapt}}	&	\multirow{2}{*}{0.8023}	&	\multirow{2}{*}{0.8789}	\\ &&&\\
    	&	\multirow{2}{*}{LCF-ATEPC \citep{yang2019multi}}	&	\multirow{2}{*}{0.8229}	&	\multirow{2}{*}{0.9018}	\\ &&&\\
        \bottomrule
    \end{tabular}
    \end{adjustbox}
    \caption{Development of the SOTA Accuracy for the aspect term polarity task \citep[SemEval-2014;][]{semeval2014}. Selected models were picked from \url{https://paperswithcode.com/sota/aspect-based-sentiment-analysis-on-semeval}.}
    \label{tab:semeval14-sb2}
\end{table}

\noindent Clearly more related, but unfortunately also less used, are the subtasks SB3 (\textit{aspect category extraction}; comparable to Subtask C1) and SB4 (\textit{aspect category polarity}; comparable to Subtask C2) from SemEval-2014.\footnote{Since the data sets (\textit{Restaurants} and \textit{Laptops}) have been further developed for SemEval-2015 and SemEval-2016, subtasks SB3 and SB4 are revisited under the names Slot 1 and Slot 3 for the in-domain ABSA in SemEval-2015. Slot 2 from SemEval-2015 aims at OTE and thus corresponds to Subtask D from GermEval17. For SemEval-2016 the same task names as in 2015 were used, subdivided into Subtask 1 (\textit{sentence-level ABSA}) and Subtask 2 (\textit{text-level ABSA)}.} Limitations with respect to comparability arise from the different numbers of categories: Subtask SB4 only exhibits five aspect categories (as opposed to 20 categories for GermEval17) which leads to an easier classification problem and is reflected in the already pretty high scores of the 2014 baselines. Table \ref{tab:semeval14-sb3-4} shows the performance of the best model from 2014 as well as performance of subsequent (pre-BERT and BERT-based) models for subtasks SB3 and SB4.

\begin{table}[ht]
    \centering
    \begin{adjustbox}{width=.49\textwidth}
    \begin{tabular}{|c|l||c|c|}
    \toprule
        &                         & \multicolumn{2}{c|}{Restaurants}\\
        & \textbf{Language model} & SB3 &  SB4  \\
        \midrule
        \multirow{3}{*}{\rotatebox[origin=c]{90}{pre-BERT}} & Best model SemEval-2014 & \multirow{2}{*}{0.8857} & \multirow{2}{*}{0.8292} \\
        & \citep{semeval2014} &&\\
    	&	\multirow{2}{*}{ATAE-LSTM \citep{wang2016attention}} &	\multirow{2}{*}{----}	&	\multirow{2}{*}{0.840}	\\  &&&\\
	\midrule
	    &	\multirow{2}{*}{BERT-pair \citep{sun2019utilizing}}	&	\multirow{2}{*}{0.9218}	&	\multirow{2}{*}{0.899}	\\ \multirow{3}{*}{\rotatebox[origin=c]{90}{BERT-based}}&&&\\
	    &	\multirow{2}{*}{CG-BERT \citep{wu2020context}}	&	\multirow{2}{*}{0.9162\textsuperscript{\textdagger}}	&	\multirow{2}{*}{0.901\textsuperscript{\textdagger}}	\\  &&&\\
    	&	\multirow{2}{*}{QACG-BERT \citep{wu2020context}}	&	\multirow{2}{*}{0.9264}	&	\multirow{2}{*}{0.904\textsuperscript{\textdagger}}	\\ &&&\\
        \bottomrule
    \end{tabular}
    \end{adjustbox}
    \caption{Development of the SOTA F1 score (SB3) and Accuracy (SB4) for the aspect category extraction/polarity task \citep[SemEval-2014;][]{semeval2014}.
    \newline \textsuperscript{\textdagger}\textit{Additional auxiliary sentences were used.}}
    \label{tab:semeval14-sb3-4}
\end{table}

\noindent In contrast to what can be observed for SB2, in this case, the performance increase on SB4 caused by the introduction of BERT seems to be kind of striking. While the ATAE-LSTM \citep{wang2016attention} only slightly increased the performance compared to 2014, the BERT-based models led to a jump of more than 6 percentage points. So when taking into account the potential room for improvement ($0.16$ for SB4 vs. $0.60$ for C2), the improvements \textit{relative} to the potential ($0.06 / 0.16$ for SB4 vs. $0.23 / 0.60$ for C2) are quite similar.

Another issue is that (partly) highly specialized (T)ABSA architectures were used for improving the SOTA on the SemEval-2014 tasks, while we "only" applied standard pre-trained German BERT models without any task-specific modifications or extensions. This leaves room for further improvements on this task on German data which should be an objective for future research.

\section{Conclusion}
\label{sec:conc}

As one would have hoped, all the state-of-the art pre-trained language models clearly outperform all the models from 2017, proving the power of transfer learning also for German ABSA. Throughout the presented analyses, the models always achieve similar results between the synchronic and the diachronic test sets, indicating temporal robustness for the models. Nonetheless, the diachronic data was collected \textit{only} half a year after the main data. It would be interesting to see whether the trained models would return similar predictions on data collected a couple of years later.

The uncased German BERT-BASE model by \texttt{dbmdz} achieves the best results across all subtasks. Since \citet{ronnqvist-etal-2019-multilingual} showed that monolingual BERT models often outperform the multilingual models for a variety of tasks, one might have already suspected that a monolingual German BERT performs best across the performed tasks. 
It may not seem evident at first that an uncased language model ends up as the best performing model since, e.g. in Sentiment Analysis, capitalized letters might be an indicator for polarity. In addition, since nouns and beginnings of sentences always start with a capital letter in German, one might assume that lower-casing the whole text changes the meaning of some words and thus confuses the language model. Nevertheless, the GermEval17 documents are very noisy since they were retrieved from social media. That means that the data contains many misspellings, grammar and expression mistakes, dialect, and colloquial language. For this reason, already some participating teams in 2017 pursued an elaborate pre-processing on the text data in order to eliminate some noise \citep{fhdo, IDS_IUCL, PotTS}. Among other things, \citet{fhdo} transformed the text to lower-case and replaced, for example, "S-Bahn" and "S Bahn" with "sbahn".
We suppose that in this case, lower-casing the texts improves the data quality by eliminating some of the noise and acts as a sort of regularization. As a result, the uncased models potentially generalize better than the cased models. The findings from \citet{mayhew2019ner}, who compare cased and uncased pre-trained models on social media data for NER, corroborate this hypothesis. 

\bibliographystyle{acl_natbib}
\bibliography{literature}

\begin{thebibliography}{}

\bibitem[Attia et~al., 2018]{attia-etal-2018-multilingual}
Attia, M., Samih, Y., Elkahky, A., and Kallmeyer, L. (2018).
\newblock Multilingual multi-class sentiment classification using convolutional
  neural networks.
\newblock In {\em Proceedings of the Eleventh International Conference on
  Language Resources and Evaluation ({LREC} 2018)}, Miyazaki, Japan. European
  Language Resources Association (ELRA).

\bibitem[Barriere and Balahur, 2020]{barriere2020improving}
Barriere, V. and Balahur, A. (2020).
\newblock Improving sentiment analysis over non-{E}nglish tweets using
  multilingual transformers and automatic translation for data-augmentation.
\newblock In {\em Proceedings of the 28th International Conference on
  Computational Linguistics}, pages 266--271, Barcelona, Spain (Online).
  International Committee on Computational Linguistics.

\bibitem[Behdenna et~al., 2018]{behdenna2018document}
Behdenna, S., Barigou, F., and Belalem, G. (2018).
\newblock Document level sentiment analysis: A survey.
\newblock {\em EAI Endorsed Transactions on Context-aware Systems and
  Applications}, 4:154339.

\bibitem[Biesialska et~al., 2020]{biesialska2020sentiment}
Biesialska, K., Biesialska, M., and Rybinski, H. (2020).
\newblock Sentiment analysis with contextual embeddings and self-attention.
\newblock {\em arXiv preprint arXiv:2003.05574}.

\bibitem[Bojanowski et~al., 2017]{bojanowski2017enriching}
Bojanowski, P., Grave, E., Joulin, A., and Mikolov, T. (2017).
\newblock Enriching word vectors with subword information.
\newblock {\em Transactions of the Association for Computational Linguistics},
  5:135--146.

\bibitem[Cho et~al., 2014]{cho2014learning}
Cho, K., Van~Merri{\"e}nboer, B., Gulcehre, C., Bahdanau, D., Bougares, F.,
  Schwenk, H., and Bengio, Y. (2014).
\newblock Learning phrase representations using rnn encoder-decoder for
  statistical machine translation.
\newblock {\em arXiv preprint arXiv:1406.1078}.

\bibitem[Cieliebak et~al., 2017]{cieliebak2017}
Cieliebak, M., Deriu, J.~M., Egger, D., and Uzdilli, F. (2017).
\newblock A {T}witter corpus and benchmark resources for {G}erman sentiment
  analysis.
\newblock In {\em Proceedings of the Fifth International Workshop on Natural
  Language Processing for Social Media}, pages 45--51, Valencia, Spain.
  Association for Computational Linguistics.

\bibitem[Devlin et~al., 2019]{devlin2019bert}
Devlin, J., Chang, M.-W., Lee, K., and Toutanova, K. (2019).
\newblock {BERT: Pre-training of Deep Bidirectional Transformers for Language
  Understanding}.
\newblock In {\em Proceedings of the 2019 Conference of the North {A}merican
  Chapter of the Association for Computational Linguistics: Human Language
  Technologies, Volume 1 (Long and Short Papers)}, pages 4171--4186,
  Minneapolis, Minnesota. Association for Computational Linguistics.

\bibitem[Espl{\`a}-Gomis et~al., 2019]{paracrawl}
Espl{\`a}-Gomis, M., Forcada, M., Ram{\'i}rez-S{\'a}nchez, G., and Hoang, H.~T.
  (2019).
\newblock {ParaCrawl: Web-scale parallel corpora for the languages of the EU}.
\newblock In {\em MTSummit}.

\bibitem[Guhr et~al., 2020]{guhr2020training}
Guhr, O., Schumann, A.-K., Bahrmann, F., and Böhme, H.-J. (2020).
\newblock {Training a Broad-Coverage German Sentiment Classification Model for
  Dialog Systems}.
\newblock In {\em {Proceedings of the 12th Conference on Language Resources and
  Evaluation (LREC 2020)}}, pages 1627--1632, Marseille, France.

\bibitem[Haddow, 2018]{newscrawl}
Haddow, B. (2018).
\newblock {News Crawl Corpus}.

\bibitem[{Hinton et~al.(2015)Hinton, Vinyals, and Dean}]{hinton2015distilling}
Geoffrey Hinton, Oriol Vinyals, and Jeff Dean. 2015.
\newblock Distilling the knowledge in a neural network.
\newblock \emph{arXiv preprint arXiv:1503.02531}.

\bibitem[Hoang et~al., 2019]{hoang-etal-2019-aspect}
Hoang, M., Bihorac, O.~A., and Rouces, J. (2019).
\newblock Aspect-based sentiment analysis using {BERT}.
\newblock In {\em Proceedings of the 22nd Nordic Conference on Computational
  Linguistics}, pages 187--196, Turku, Finland. Link{\"o}ping University
  Electronic Press.

\bibitem[Hochreiter and Schmidhuber, 1997]{hochreiter1997long}
Hochreiter, S. and Schmidhuber, J. (1997).
\newblock Long short-term memory.
\newblock {\em Neural computation}, 9(8):1735--1780.

\bibitem[Hövelmann and Friedrich, 2017]{fhdo}
Hövelmann, L. and Friedrich, C.~M. (2017).
\newblock {Fasttext and Gradient Boosted Trees at GermEval-2017 Tasks on
  Relevance Classification and Document-level Polarity}.
\newblock In {\em {Proceedings of the GermEval 2017 – Shared Task on
  Aspect-based Sentiment in Social Media Customer Feedback}}, Berlin, Germany.

\bibitem[Karimi et~al., 2020]{karimi2020adversarial}
Karimi, A., Rossi, L., and Prati, A. (2020).
\newblock Adversarial training for aspect-based sentiment analysis with bert.

\bibitem[Lee et~al., 2017]{UKP_Lab_TUDA}
Lee, J.-U., Eger, S., Daxenberger, J., and Gurevych, I. (2017).
\newblock {UKP TU-DA at GermEval 2017: Deep Learning for Aspect Based Sentiment
  Detection}.
\newblock In {\em {Proceedings of the GermEval 2017 – Shared Task on
  Aspect-based Sentiment in Social Media Customer Feedback}}, Berlin, Germany.

\bibitem[Li et~al., 2018]{li-etal-2018-hierarchical}
Li, L., Liu, Y., and Zhou, A. (2018).
\newblock Hierarchical attention based position-aware network for aspect-level
  sentiment analysis.
\newblock In {\em Proceedings of the 22nd Conference on Computational Natural
  Language Learning}, pages 181--189, Brussels, Belgium. Association for
  Computational Linguistics.

\bibitem[Li et~al., 2019]{li2019exploiting}
Li, X., Bing, L., Zhang, W., and Lam, W. (2019).
\newblock Exploiting {BERT} for end-to-end aspect-based sentiment analysis.
\newblock In {\em Proceedings of the 5th Workshop on Noisy User-generated Text
  (W-NUT 2019)}, pages 34--41, Hong Kong, China. Association for Computational
  Linguistics.

\bibitem[Lison and Tiedemann, 2016]{opensubtitles}
Lison, P. and Tiedemann, J. (2016).
\newblock {OpenSubtitles2016: Extracting Large Parallel Corpora from Movie and
  TV Subtitles}.
\newblock In {\em Proceedings of the 10th International Conference on Language
  Resources and Evaluation (LREC 2016)}.

\bibitem[Mayhew et~al., 2019]{mayhew2019ner}
Mayhew, S., Tsygankova, T., and Roth, D. (2019).
\newblock ner and pos when nothing is capitalized.
\newblock In {\em Proceedings of the 2019 Conference on Empirical Methods in
  Natural Language Processingand the 9th International Joint Conference on
  Natural Language Processing}, pages 6256--6261, Hong Kong, China. Association
  for Computational Linguistics.

\bibitem[McCann et~al., 2017]{mccann2017learned}
McCann, B., Bradbury, J., Xiong, C., and Socher, R. (2017).
\newblock Learned in translation: Contextualized word vectors.
\newblock In Guyon, I., Luxburg, U.~V., Bengio, S., Wallach, H., Fergus, R.,
  Vishwanathan, S., and Garnett, R., editors, {\em Advances in Neural
  Information Processing Systems}, volume~30, pages 6294--6305. Curran
  Associates, Inc.

\bibitem[Mishra et~al., 2017]{im_sing}
Mishra, P., Mujadia, V., and Lanka, S. (2017).
\newblock {GermEval 2017: Sequence based Models for Customer Feedback
  Analysis}.
\newblock In {\em {Proceedings of the GermEval 2017 – Shared Task on
  Aspect-based Sentiment in Social Media Customer Feedback}}, Berlin, Germany.

\bibitem[Ortiz~Su{\'a}rez et~al., 2019]{commoncrawl}
Ortiz~Su{\'a}rez, P.~J., Sagot, B., and Romary, L. (2019).
\newblock {Asynchronous Pipeline for Processing Huge Corpora on Medium to Low
  Resource Infrastructures}.
\newblock In Ba{\'n}ski, P., Barbaresi, A., Biber, H., Breiteneder, E.,
  Clematide, S., Kupietz, M., L{\"u}ngen, H., and Iliadi, C., editors, {\em
  {7th Workshop on the Challenges in the Management of Large Corpora
  (CMLC-7)}}, Cardiff, United Kingdom. {Leibniz-Institut f{\"u}r Deutsche
  Sprache}.

\bibitem[Ostendorff et~al., 2020]{openlegaldata}
Ostendorff, M., Blume, T., and Ostendorff, S. (2020).
\newblock {Towards an Open Platform for Legal Information}.
\newblock In {\em Proceedings of the ACM/IEEE Joint Conference on Digital
  Libraries in 2020}, JCDL '20, pages 385–--388, New York, NY, USA.
  Association for Computing Machinery.

\bibitem[Pennington et~al., 2014]{pennington2014glove}
Pennington, J., Socher, R., and Manning, C. (2014).
\newblock Glove: Global vectors for word representation.
\newblock In {\em Proceedings of the 2014 conference on empirical methods in
  natural language processing (EMNLP)}, pages 1532--1543.

\bibitem[Peters et~al., 2018]{peters2018deep}
Peters, M.~E., Neumann, M., Iyyer, M., Gardner, M., Clark, C., Lee, K., and
  Zettlemoyer, L. (2018).
\newblock Deep contextualized word representations.
\newblock {\em arXiv preprint arXiv:1802.05365}.

\bibitem[Pontiki et~al., 2016]{semeval2016}
Pontiki, M., Galanis, D., Papageorgiou, H., Androutsopoulos, I., Manandhar, S.,
  AL-Smadi, M., Al-Ayyoub, M., Zhao, Y., Qin, B., de~clercq, O., Hoste, V.,
  Apidianaki, M., Tannier, X., Loukachevitch, N., Kotelnikov, E., Bel, N.,
  Zafra, S.~M., and Eryiğit, G. (2016).
\newblock Semeval-2016 task 5: Aspect based sentiment analysis.
\newblock In {\em Proceedings of the 10th International Workshop on Semantic
  Evaluation (SemEval-2016)}, pages 19--30.

\bibitem[Pontiki et~al., 2015]{semeval2015}
Pontiki, M., Galanis, D., Papageorgiou, H., Manandhar, S., and Androutsopoulos,
  I. (2015).
\newblock {S}em{E}val-2015 task 12: Aspect based sentiment analysis.
\newblock In {\em Proceedings of the 9th International Workshop on Semantic
  Evaluation ({S}em{E}val 2015)}, pages 486--495, Denver, Colorado. Association
  for Computational Linguistics.

\bibitem[Pontiki et~al., 2014]{semeval2014}
Pontiki, M., Galanis, D., Pavlopoulos, J., Papageorgiou, H., Androutsopoulos,
  I., and Manandhar, S. (2014).
\newblock {S}em{E}val-2014 task 4: Aspect based sentiment analysis.
\newblock In {\em Proceedings of the 8th International Workshop on Semantic
  Evaluation ({S}em{E}val 2014)}, pages 27--35, Dublin, Ireland. Association
  for Computational Linguistics.

\bibitem[Rietzler et~al., 2020]{rietzler-etal-2020-adapt}
Rietzler, A., Stabinger, S., Opitz, P., and Engl, S. (2020).
\newblock Adapt or get left behind: Domain adaptation through {BERT} language
  model finetuning for aspect-target sentiment classification.
\newblock In {\em Proceedings of the 12th Language Resources and Evaluation
  Conference}, pages 4933--4941, Marseille, France. European Language Resources
  Association.

\bibitem[R{\"o}nnqvist et~al., 2019]{ronnqvist-etal-2019-multilingual}
R{\"o}nnqvist, S., Kanerva, J., Salakoski, T., and Ginter, F. (2019).
\newblock {Is Multilingual BERT Fluent in Language Generation?}
\newblock In {\em {Proceedings of the First NLPL Workshop on Deep Learning for
  Natural Language Processing}}, pages 29--36, Turku, Finland. Link{\"o}ping
  University Electronic Press.

\bibitem[Ruppert et~al., 2017]{LT-ABSA}
Ruppert, E., Kumar, A., and Biemann, C. (2017).
\newblock {LT-ABSA: An Extensible Open-Source System for Document-Level and
  Aspect-Based Sentiment Analysis}.
\newblock In {\em {Proceedings of the GermEval 2017 – Shared Task on
  Aspect-based Sentiment in Social Media Customer Feedback}}, Berlin, Germany.

\bibitem[{Sanh et~al.(2019)Sanh, Debut, Chaumond, and
  Wolf}]{sanh2019distilbert}
Victor Sanh, Lysandre Debut, Julien Chaumond, and Thomas Wolf. 2019.
\newblock Distilbert, a distilled version of bert: smaller, faster, cheaper and
  lighter.
\newblock \emph{arXiv preprint arXiv:1910.01108}.

\bibitem[Sayyed et~al., 2017]{IDS_IUCL}
Sayyed, Z.~A., Dakota, D., and Kübler, S. (2017).
\newblock {IDS-IUCL: Investigating Feature Selection and Oversampling for
  GermEval 2017}.
\newblock In {\em {Proceedings of the GermEval 2017 – Shared Task on
  Aspect-based Sentiment in Social Media Customer Feedback}}, Berlin, Germany.

\bibitem[Schmitt et~al., 2018]{schmitt-etal-2018-joint}
Schmitt, M., Steinheber, S., Schreiber, K., and Roth, B. (2018).
\newblock Joint aspect and polarity classification for aspect-based sentiment
  analysis with end-to-end neural networks.
\newblock In {\em Proceedings of the 2018 Conference on Empirical Methods in
  Natural Language Processing}, pages 1109--1114, Brussels, Belgium.
  Association for Computational Linguistics.

\bibitem[Sidarenka, 2017]{PotTS}
Sidarenka, U. (2017).
\newblock {PotTS at GermEval-2017 Task B: Document-Level Polarity Detection
  Using Hand-Crafted SVM and Deep Bidirectional LSTM Network}.
\newblock In {\em {Proceedings of the GermEval 2017 – Shared Task on
  Aspect-based Sentiment in Social Media Customer Feedback}}, Berlin, Germany.

\bibitem[Skadi{\c{n}}{\v{s}} et~al., 2014]{eubookshop}
Skadi{\c{n}}{\v{s}}, R., Tiedemann, J., Rozis, R., and Deksne, D. (2014).
\newblock {Billions of Parallel Words for Free: Building and Using the {EU}
  Bookshop Corpus}.
\newblock In {\em Proceedings of the 9th International Conference on Language
  Resources and Evaluation ({LREC} 2014)}, pages 1850--1855, Reykjavik,
  Iceland. European Language Resources Association (ELRA).

\bibitem[Song et~al., 2019]{song2019attentional}
Song, Y., Wang, J., Jiang, T., Liu, Z., and Rao, Y. (2019).
\newblock Attentional encoder network for targeted sentiment classification.
\newblock {\em arXiv preprint arXiv:1902.09314}.

\bibitem[Sun et~al., 2019]{sun2019utilizing}
Sun, C., Huang, L., and Qiu, X. (2019).
\newblock Utilizing {BERT} for aspect-based sentiment analysis via constructing
  auxiliary sentence.
\newblock In {\em Proceedings of the 2019 Conference of the North {A}merican
  Chapter of the Association for Computational Linguistics: Human Language
  Technologies, Volume 1 (Long and Short Papers)}, pages 380--385, Minneapolis,
  Minnesota. Association for Computational Linguistics.

\bibitem[Tang et~al., 2016]{tang2016aspect}
Tang, D., Qin, B., and Liu, T. (2016).
\newblock Aspect level sentiment classification with deep memory network.
\newblock {\em arXiv preprint arXiv:1605.08900}.

\bibitem[Tao and Fang, 2020]{tao2020toward}
Tao, J. and Fang, X. (2020).
\newblock Toward multi-label sentiment analysis: a transfer learning based
  approach.
\newblock {\em Journal of Big Data}, 7:1.

\bibitem[Vaswani et~al., 2017]{vaswani2017attention}
Vaswani, A., Shazeer, N., Parmar, N., Uszkoreit, J., Jones, L., Gomez, A.~N.,
  Kaiser, L., and Polosukhin, I. (2017).
\newblock {Attention Is All You Need}.
\newblock In {\em 31st Conference on Neural Information Processing Systems
  (NIPS 2017)}, Long Beach, California, USA.

\bibitem[Wang et~al., 2019]{wang2019superglue}
Wang, A., Pruksachatkun, Y., Nangia, N., Singh, A., Michael, J., Hill, F.,
  Levy, O., and Bowman, S. (2019).
\newblock Superglue: A stickier benchmark for general-purpose language
  understanding systems.
\newblock In {\em Advances in neural information processing systems}, pages
  3266--3280.

\bibitem[Wang et~al., 2018]{wang2018glue}
Wang, A., Singh, A., Michael, J., Hill, F., Levy, O., and Bowman, S.~R. (2018).
\newblock Glue: A multi-task benchmark and analysis platform for natural
  language understanding.
\newblock {\em arXiv preprint arXiv:1804.07461}.

\bibitem[Wang et~al., 2016]{wang2016attention}
Wang, Y., Huang, M., Zhu, X., and Zhao, L. (2016).
\newblock Attention-based lstm for aspect-level sentiment classification.
\newblock In {\em Proceedings of the 2016 conference on empirical methods in
  natural language processing}, pages 606--615.

\bibitem[Wojatzki et~al., 2017]{wojatzki2017germeval}
Wojatzki, M., Ruppert, E., Holschneider, S., Zesch, T., and Biemann, C. (2017).
\newblock {GermEval 2017: Shared Task on Aspect-based Sentiment in Social Media
  Customer Feedback}.
\newblock In {\em {Proceedings of the GermEval 2017 – Shared Task on
  Aspect-based Sentiment in Social Media Customer Feedback}}, pages 1--12,
  Berlin, Germany.

\bibitem[Wolf et~al., 2020]{Wolf2020HuggingFace}
Wolf, T., Debut, L., Sanh, V., Chaumond, J., Delangue, C., Moi, A., Cistac, P.,
  Rault, T., Louf, R., Funtowicz, M., Davison, J., Shleifer, S., von Platen,
  P., Ma, C., Jernite, Y., Plu, J., Xu, C., Scao, T.~L., Gugger, S., Drame, M.,
  Lhoest, Q., and Rush, A.~M. (2020).
\newblock {Transformers: State-of-the-Art Natural Language Processing}.
\newblock In {\em Proceedings of the 2020 Conference on Empirical Methods in
  Natural Language Processing: System Demonstrations}, pages 38--45, Online.
  Association for Computational Linguistics.

\bibitem[Wu and Ong, 2020]{wu2020context}
Wu, Z. and Ong, D.~C. (2020).
\newblock Context-guided bert for targeted aspect-based sentiment analysis.
\newblock {\em arXiv preprint arXiv:2010.07523}.

\bibitem[Xu et~al., 2019]{xu2019bert}
Xu, H., Liu, B., Shu, L., and Yu, P.~S. (2019).
\newblock Bert post-training for review reading comprehension and aspect-based
  sentiment analysis.

\bibitem[Yang et~al., 2019]{yang2019multi}
Yang, H., Zeng, B., Yang, J., Song, Y., and Xu, R. (2019).
\newblock A multi-task learning model for chinese-oriented aspect polarity
  classification and aspect term extraction.
\newblock {\em arXiv preprint arXiv:1912.07976}.

\end{thebibliography}

\section*{Appendix}

\appendix

\section{Detailed results (per category) for Subtask C}
\label{a:details-c}

It may be interesting to have a more detailed look at the model performance for this subtask because of the high number of classes and their skewed distribution by investigating the performance on category-level.
Table \ref{tab:detail_C1} shows the performance of the uncased German BERT-BASE model by \texttt{dbmdz} per test set for Subtask C1. The support indicates the number of appearances, which are also displayed in Table \ref{tab:distribution_aspects} in this case. Seven categories are summarized in \textit{Rest} because they have an F1 score of 0 for both test sets, i.e. the model is not able to correctly identify any of these seven aspects appearing in the test data. The table is sorted by the score on the synchronic test set.

\begin{table}[ht]
    \centering
    \begin{adjustbox}{width=.49\textwidth}
    \begin{tabular}{|l|rr||rr|}
    \hline
        & \multicolumn{2}{c||}{$\textbf{test}_{syn}$} & \multicolumn{2}{c|}{$\textbf{test}_{dia}$} \\
        \textbf{Aspect Category} & \textbf{Score} & \textbf{Support} & \textbf{Score} & \textbf{Support} \\
        \hline
Allgemein	&	0.854	&	1,398	&	0.877	&	1,024	\\
Sonstige Unregelmäßigkeiten	&	0.782	&	224	&	0.785	&	164	\\
Connectivity	&	0.750	&	36	&	0.838	&	73	\\
Zugfahrt	&	0.678	&	241	&	0.687	&	184	\\
Auslastung und Platzangebot	&	0.645	&	35	&	0.667	&	20	\\
Sicherheit	&	0.602	&	84	&	0.639	&	42	\\
Atmosphäre	&	0.600	&	148	&	0.532	&	53	\\
Barrierefreiheit	&	0.500	&	9	&	0	&	2	\\
Ticketkauf	&	0.481	&	95	&	0.506	&	48	\\
Service und Kundenbetreuung	&	0.476	&	63	&	0.417	&	27	\\
DB App und Website	&	0.455	&	28	&	0.563	&	18	\\
Informationen	&	0.329	&	58	&	0.464	&	35	\\
Komfort und Ausstattung	&	0.286	&	24	&	0	&	11	\\
\textit{Rest}	&	\textit{0}	&	\textit{24}	&	\textit{0}	&	\textit{20}	\\
    \hline
    \end{tabular}
    \end{adjustbox}
    \caption{Micro-averaged F1 scores and support by aspect category (Subtask C1). Seven categories are summarized in \textit{Rest} and show each a score of 0.}
    \label{tab:detail_C1}
\end{table}

\noindent The F1 scores for \texttt{Allgemein} (\textit{General}), \texttt{Sonstige Unregelmäßigkeiten} (\textit{Other irregularities}) and \texttt{Connectivity} are the highest. 13 categories, mostly similar between the two test sets, show a positive F1 score on at least one of the two test sets. For the categories subsumed under \textit{Rest}, the model was not able to learn how to correctly identify these categories.

Subtask C2 exhibits a similar distribution of the true labels, with the \textit{Aspect+Sentiment} category \texttt{Allgemein:neutral} as majority class. Over 50\% of the true labels belong to this class. Table \ref{tab:detail_C2} shows that only 12 out of 60 labels can be detected by the model (see Table \ref{tab:detail_C2}).

\begin{table}[ht]
    \centering
    \begin{adjustbox}{width=.49\textwidth}
    \begin{tabular}{|l|rr||rr|}
    \hline
        & \multicolumn{2}{c||}{$\textbf{test}_{syn}$} & \multicolumn{2}{c|}{$\textbf{test}_{dia}$} \\
        \textbf{Aspect+Sentiment Category} & \textbf{Score} & \textbf{Support} & \textbf{Score} & \textbf{Support} \\
        \hline
Allgemein:neutral	&	0.804	&	1,108	&	0.832	&	913	\\
Sonstige Unregelmäßigkeiten:negative	&	0.782	&	221	&	0.793	&	159	\\
Zugfahrt:negative	&	0.645	&	197	&	0.725	&	149	\\
Sicherheit:negative	&	0.640	&	78	&	0.585	&	39	\\
Allgemein:negative	&	0.582	&	258	&	0.333	&	80	\\
Atmosphäre:negative	&	0.569	&	126	&	0.447	&	39	\\
Connectivity:negative	&	0.400	&	20	&	0.291	&	46	\\
Ticketkauf:negative	&	0.364	&	42	&	0.298	&	34	\\
Auslastung und Platzangebot:negative	&	0.350	&	31	&	0.211	&	17	\\
Allgemein:positive	&	0.214	&	41	&	0.690	&	33	\\
Zugfahrt:positive	&	0.154	&	34	&	0	&	34	\\
Service und Kundenbetreuung:negative	&	0.146	&	36	&	0.174	&	21	\\
\textit{Rest}	&	\textit{0}	&	\textit{343}	&	\textit{0}	&	\textit{180}	\\
        \hline
    \end{tabular}
    \end{adjustbox}
    \caption{Micro-averaged F1 scores and support by \textit{Aspect+Sentiment} category (Subtask C2). 48 categories are summarized in \textit{Rest} and show each a score of 0.}
    \label{tab:detail_C2}
\end{table}

All the aspect categories displayed in Table \ref{tab:detail_C2} are also visible in Table \ref{tab:detail_C1} and most of them have negative sentiment. \texttt{Allgemein:neutral} and \texttt{Sonstige Unregelmäßigkeiten:negative} show the highest scores. Again, we assume that here, 48 categories could not be identified due to data sparsity. However, having this in mind, the model achieves a relatively high overall performance for both, Subtask C1 and C2 (cf. Tab. \ref{tab:myresultsC1} and Tab. \ref{tab:myresultsC2}). This is mainly owed to the high score of the majority classes \texttt{Allgemein} and \texttt{Allgemein:neutral}, respectively, because the micro F1 score puts a lot of weight on majority classes. It might be interesting whether the classification of the rare categories can be improved by balancing the data. We experimented with removing general categories such as \texttt{Allgemein}, \texttt{Allgemein:neutral} or documents with sentiment \texttt{neutral} since these are usually less interesting for a company. We observe a large drop in the overall F1 score which is attributed to the absence of the strong majority class and the resulting data loss. Indeed, the classification for some single categories could be improved, but the rare categories could still not be identified by the model.

\section{Detailed results (per category) for Subtask D}
\label{a:details-d}

Similar as for Subtask C, the results for the best model are investigated in more detail. Table \ref{tab:detail_D1} gives the detailed classification report for the uncased German BERT-BASE model with CRF layer on Subtask D1. Only entities that were correctly detected at least once are displayed. The table is sorted by the score on the synchronic test set.
The classification report for Subtask D2 is displayed analogously in Table \ref{tab:detail_D2}.

\begin{table}[ht]
    \centering
    \begin{adjustbox}{width=.49\textwidth}
    \begin{tabular}{|l|rr||rr|}
    \hline
    & \multicolumn{2}{c||}{$\textbf{test}_{syn}$} & \multicolumn{2}{c|}{$\textbf{test}_{dia}$} \\
    \textbf{Category} & \textbf{Score} & \textbf{Support} & \textbf{Score} & \textbf{Support} \\
    \hline
Zugfahrt:negative	&	0.702	&	622	&	0.729	&	495	\\
Sonstige Unregelmäßigkeiten:negative	&	0.681	&	693	&	0.581	&	484	\\
Sicherheit:negative	&	0.604	&	337	&	0.457	&	122	\\
Connectivity:negative	&	0.598	&	56	&	0.620	&	109	\\
Barrierefreiheit:negative	&	0.595	&	14	&	0	&	3	\\
Auslastung und Platzangebot:negative	&	0.579	&	66	&	0.447	&	31	\\
Connectivity:positive	&	0.571	&	26	&	0.555	&	60	\\
Allgemein:negative	&	0.545	&	807	&	0.343	&	139	\\
Atmosphäre:negative	&	0.500	&	403	&	0.337	&	164	\\
Ticketkauf:negative	&	0.383	&	96	&	0.583	&	74	\\
Ticketkauf:positive	&	0.368	&	59	&	0	&	13	\\
Komfort und Ausstattung:negative	&	0.357	&	24	&	0	&	16	\\
Atmosphäre:neutral	&	0.348	&	40	&	0.111	&	14	\\
Service und Kundenbetreuung:negative	&	0.323	&	74	&	0.286	&	31	\\
Informationen:negative	&	0.301	&	68	&	0.505	&	46	\\
Zugfahrt:positive	&	0.276	&	62	&	0.343	&	83	\\
DB App und Website:negative	&	0.232	&	39	&	0.375	&	33	\\
DB App und Website:neutral	&	0.188	&	23	&	0	&	11	\\
Sonstige Unregelmäßigkeiten:neutral	&	0.179	&	13	&	0.222	&	2	\\
Allgemein:positive	&	0.157	&	86	&	0.586	&	92	\\
Service und Kundenbetreuung:positive	&	0.115	&	23	&	0	&	5	\\
Atmosphäre:positive	&	0.105	&	26	&	0	&	15	\\
Ticketkauf:neutral	&	0.040	&	144	&	0.222	&	25	\\
Connectivity:neutral	&	0	&	11	&	0.211	&	15	\\
Toiletten:negative	&	0	&	15	&	0.160	&	23	\\
\textit{Rest}	&	\textit{0}	&	\textit{355}	&	\textit{0}	&	\textit{115}	\\
\hline
    \end{tabular}
    \end{adjustbox}
    \caption{Micro-averaged F1 scores and support by \textit{Aspect+Sentiment} entity with exact match (Subtask D1). 35 categories are summarized in \textit{Rest}, each of them exhibiting a score of 0.}
    \label{tab:detail_D1}
\end{table}

\noindent For Subtask D1, the model returns a positive score on 25 entity categories on at least one of the two test sets. The category \texttt{Zugfahrt:negative} can be classified best on both test sets, followed by \texttt{Sonstige Unregelmäßigkeiten:negative} and \texttt{Sicherheit:negative} for the synchronic test set and by \texttt{Connectivity:negative} and \texttt{Allgemein:positive} for the diachronic set. Visibly, the scores between the two test sets differ more here than in the classification report of the previous task.

\begin{table}[ht]
    \centering
    \begin{adjustbox}{width=.49\textwidth}
    \begin{tabular}{|l|rr||rr|}
    \hline
    & \multicolumn{2}{c||}{$\textbf{test}_{syn}$} & \multicolumn{2}{c|}{$\textbf{test}_{dia}$} \\
    \textbf{Category} & \textbf{Score} & \textbf{Support} & \textbf{Score} & \textbf{Support} \\
    \hline
Zugfahrt:negative	&	0.708	&	622	&	0.739	&	495	\\
Sonstige Unregelmäßigkeiten:negative	&	0.697	&	693	&	0.617	&	484	\\
Sicherheit:negative	&	0.607	&	337	&	0.475	&	122	\\
Connectivity:negative	&	0.598	&	56	&	0.620	&	109	\\
Barrierefreiheit:negative	&	0.595	&	14	&	0	&	3	\\
Auslastung und Platzangebot:negative	&	0.579	&	66	&	0.447	&	31	\\
Connectivity:positive	&	0.571	&	26	&	0.555	&	60	\\
Allgemein:negative	&	0.561	&	807	&	0.363	&	139	\\
Atmosphäre:negative	&	0.505	&	403	&	0.358	&	164	\\
Ticketkauf:negative	&	0.383	&	96	&	0.583	&	74	\\
Ticketkauf:positive	&	0.368	&	59	&	0	&	13	\\
Komfort und Ausstattung:negative	&	0.357	&	24	&	0	&	16	\\
Atmosphäre:neutral	&	0.348	&	40	&	0.111	&	14	\\
Service und Kundenbetreuung:negative	&	0.323	&	74	&	0.286	&	31	\\
Informationen:negative	&	0.301	&	68	&	0.505	&	46	\\
Zugfahrt:positive	&	0.276	&	62	&	0.343	&	83	\\
DB App und Website:negative	&	0.261	&	39	&	0.406	&	33	\\
DB App und Website:neutral	&	0.188	&	23	&	0	&	11	\\
Sonstige Unregelmäßigkeiten:neutral	&	0.179	&	13	&	0.222	&	2	\\
Allgemein:positive	&	0.157	&	86	&	0.586	&	92	\\
Service und Kundenbetreuung:positive	&	0.115	&	23	&	0	&	5	\\
Atmosphäre:positive	&	0.105	&	26	&	0	&	15	\\
Ticketkauf:neutral	&	0.040	&	144	&	0.222	&	25	\\
Connectivity:neutral	&	0	&	11	&	0.211	&	15	\\
Toiletten:negative	&	0	&	15	&	0.160	&	23	\\
\textit{Rest}	&	\textit{0}	&	\textit{355}	&	\textit{0}	&	\textit{112}	\\
\hline
    \end{tabular}
    \end{adjustbox}
    \caption{Micro-averaged F1 scores and support by \textit{Aspect+Sentiment} entity with overlapping match (Subtask D2). 35 categories are summarized in \textit{Rest} and show each a score of 0.}
    \label{tab:detail_D2}
\end{table}

The report for the overlapping match (cf. Tab. \ref{tab:detail_D2}) shows slightly better results on some categories than for the exact match. The third-best score on the diachronic test data is now \texttt{Sonstige Unregelmäßigkeiten:negative}. Besides this, the top three categories per test set remain the same. 

Apart from the fact that this is a different kind of task than before, one can notice that even though the overall micro F1 scores are lower for Subtask D than for Subtask C, the model manages to successfully identify a larger variety of categories, i.e. it achieves a positive score for more categories. This is probably due to the more balanced data for Subtask D than for Subtask C2, resulting in a lower overall score and mostly higher scores per category.

\end{document}